\documentclass[sigconf]{acmart}

\usepackage{subcaption,float}
\usepackage{algorithm, algorithmic}
\usepackage{natbib} 

\AtBeginDocument{%
  }

\setcopyright{acmlicensed}
\copyrightyear{2018}
\acmYear{2018}
\acmDOI{XXXXXXX.XXXXXXX}

\acmConference[Conference acronym 'XX]{Make sure to enter the correct
  conference title from your rights confirmation emai}{June 03--05,
  2018}{Woodstock, NY}
\acmISBN{978-1-4503-XXXX-X/18/06}

\newlength\myindent
\newcommand\bindent{%
  \begingroup
  \setlength{\itemindent}{\myindent}
  \addtolength{\algorithmicindent}{\myindent}
}
\newcommand\eindent{\endgroup}

\begin{document}

\title{Relevance-aware Algorithmic Recourse}
\graphicspath{ {./images/} }

\author{Dongwhi Kim}
\affiliation{%
  \institution{University of Notre Dame}
  \country{}}
  
\email{dkim37@nd.edu}

\author{Nuno Moniz}
\affiliation{%
  \institution{University of Notre Dame}
  \country{} }
\email{nuno.moniz@nd.edu}

% \renewcommand{\shortauthors}{Anonymous}

\begin{abstract}
    As machine learning continues to gain prominence, transparency and explainability are increasingly critical. Without an understanding of these models, they can replicate and worsen human bias, adversely affecting marginalized communities. Algorithmic recourse emerges as a tool for clarifying decisions made by predictive models, providing actionable insights to alter outcomes. They answer, ``What do I have to change?'' to achieve the desired result. Despite their importance, current algorithmic recourse methods treat all domain values equally, which is unrealistic in real-world settings. In this paper, we propose a novel framework, Relevance-Aware Algorithmic Recourse (RAAR), that leverages the concept of relevance in applying algorithmic recourse to regression tasks. We conducted multiple experiments on 15 datasets to outline how relevance influences recourses. Results show that relevance contributes algorithmic recourses comparable to well-known baselines, with greater efficiency and lower relative costs.

\end{abstract}




\received{20 February 2007}
\received[revised]{12 March 2009}
\received[accepted]{5 June 2009}

\maketitle

\section{Introduction}
The landscape of machine learning applications in decision-making has exploded in the past decade, ranging from dynamic pricing algorithms in services like Uber to pivotal decisions in areas such as bail and loan approvals~\cite{loan_ml, uber_dynamic}. The sheer volume of data generation highlights the need for better ways to analyze and leverage data to provide quick and accurate decision-making~\cite{big_data, Eirinaki2016}. In this respect, machine learning has shown it can provide solutions capable of rapidly analyzing data and benefiting society, although requiring human monitoring to ensure a just experience.

Unsurprisingly, the increased adoption of machine learning has drawn attention from governments and large organizations in regulation~\cite{lse_regulate}. Evidenced by the ``right to explanation'' introduced by the EU General Data Protection Regulation (GDPR), it seeks to safeguard individuals adversely affected by semi-autonomous decisions, giving them the right to understand why they were denied~\cite{eu_gdpr}. Similarly, within the United States, the ``right to explanation'' is enforced in credit scoring, ensuring that creditors explain the key factors that influence an individual's credit score~\cite{us_ecfrFederalRegister}. Consequently, there is a growing emphasis on developing programs and analyses that provide avenues for recourse to those impacted by semi-autonomous decisions~\cite{eu_gdpr_research}. Regulation is necessary to manage the associated risks associated with AI while promoting its development.

Counterfactual explanations are excellent tools to mitigate risks, promote the ethical use of AI, and abide by regulations~\cite{counterfactual_importance}. They outline the necessary conditions to achieve a desired outcome, offering insights into complex models. In addition, they reduce the regulatory burden by providing practical recourse options and align with the GDPR's ``right to explanation"~\cite{eu_gdpr_research}.

In recent literature, multiple approaches exist for algorithmic recourse, i.e., counterfactuals, from gradient-based methods to integer programming-based approaches and probabilistic optimizations~\cite{alg_gdpr, int_prog, bo_prob}. Although different, each strategy offers nuanced perspectives and techniques to address the challenges associated with algorithmic decision-making. %If one can already receive help from many methods, why not choose just one?

\paragraph{Motivation}
Suppose a person arrives at a bank seeking a \$100,000 loan but is only allowed \$50,000. Algorithmic recourse will only be useful if it provides actionable and realistic counterfactuals, conveying how the individual may effect small changes to become eligible. As such, disregarding user preferences may lead to useless explanations. This describes the main issue with current algorithmic recourse methods. Current methods lack consideration for user preferences over the domain, i.e., relevance, and generally assume all values in a domain are equally important in obtaining an accurate prediction~\cite{relevance_nuno}. In this context, relevance is the importance of obtaining an accurate prediction from a user perspective. %For example, if a person is denied a loan of \$100,000, our current position would have a relevance of 0, \$100,000 would be 1, and the values would range from 0-1. This is necessary because instead of having a uniform domain preference where every value is equally important, we should treat these values differently because some values, like ones closer to our target outcome, should be more relevant than others.
%For example, if one wants to determine the optimal times to run wind turbines. Continuously assimilating real-time data, these turbines are designed to operate within a specific range of values~\cite{wind_turbine_opt}. Operating under excessively windy conditions may pose a risk of mechanical failure, while insufficient wind would render the operation inefficient in resource utilization. However, when employing algorithmic models in these scenarios, challenges emerge in aligning with real-world dynamics to attain an optimal outcome rather than just minimizing costs to get a certain value. These real-world constraints and objectives reinforce the need for an adaptive and relevance-aware recourse framework for practical applications. 

\paragraph{Contribution.} In this paper, we introduce a new framework, Relevance-Aware Algorithmic Recourse (RAAR) rooted in providing accurate and realistic recourses by leveraging the concept of domain relevance~\cite{relevance_nuno}. It enhances the effectiveness of algorithmic recourse in decision-making processes through counterfactual explanations that maximize one's outcomes relative to one's situation. Our results show relevance contributes to algorithmic recourses comparable to well-known baselines, with greater efficiency and lower relative costs. As a result, we can provide more actionable recourses for users, enabling them to take action more easily and still obtain their desired outcome. 

%  This paper presents a groundbreaking advancement in the field of explainable artificial intelligence, specifically focusing on algorithmic recourse, through the development of a novel framework: Relevance-Aware Algorithmic Recourse (RAAR). The RAAR framework is underpinned by the integration of domain relevance, which significantly improves the accuracy and realism of recourses within decision-making processes via contextually optimized counterfactual explanations. These enhancements result in improved outcomes for individuals, given their unique circumstances.

% Our extensive experimental evaluations demonstrate that incorporating relevance into algorithmic recourse yields outcomes that are highly competitive when compared to established baselines. Furthermore, the proposed methodology exhibits superior efficiency and reduced relative costs, making it an attractive alternative for practical applications. By bridging the gap between abstract mathematical concepts and tangible real-world impacts, this work underscores the importance of considering domain relevance in algorithmic recourse, thereby fostering more transparent, fair, and efficient AI systems.

\section{Related Works}
%\subsection{Algorithmic Recourse}
Algorithmic recourse provides actionable recourse to individuals affected by semi-autonomous decisions ~\cite{alg_rec_overview}. It answers: \textbf{how can I change my situation}~\cite{algrec_def_1, algrec_def_2}? Currently, there are several approaches that other papers have outlined to tackle this same issue. In general, they fall into these categories: level of accessibility to the predictive model (black box or gradient), the enforcement of sparsity (changing the least features) in counterfactuals, research of the underlying causal relationship, and more~\cite{alg_rec_overview}. Certain models also use feature-highlighting explanations or add actionability and plausibility constraints by setting features to be actionable and mutable, non-actionable and mutable, or non-actionable and immutable~\cite{alg_rec_principal_reasons, alg_rec_plausibility}. There has also been work to make these models robust to dataset shifts in the real world~\cite{roar}. Yet, none incorporate the use of domain preferences, i.e., relevance. In this paper, we focus our contribution in the scope of Bayesian Optimization methods.

\subsection{Bayesian Optimization}
Bayesian optimization is a global optimization problem to solve: 

% \[ x^* = argmax_{\{P^{\circ}_t(j)\}}f(x) \]
\begin{equation} 
x^*=
\max_{x \epsilon X} f(x),
\end{equation} 

\noindent
where $f(x)$ is the objective function, $X$ is the set of features, $x$ is a feature, and $x^*$ is the feature for the global maximum of the objective function~\cite{bo_tutorial}. When $f(x)$ is unknown, Bayesian optimization treats it as a black-box function, a scenario commonly encountered in algorithmic recourse~\cite{bayes_opt_info}. The methodology revolves around three primary components: objective function (black box), surrogate model (Gaussian process), and acquisition function~\cite{gp_tutorial}.
\subsubsection{Pseudo Code of the Algorithm}
\vspace*{-\baselineskip}
\begin{algorithm}[H]
\caption{Bayesian optimization pseudo code}\label{alg:cap}
\begin{algorithmic}
\STATE \textbf{function} BayesOpt({$f, M_{0}, N, A$})
\bindent
% \Function{BayesOpt}{$f, M_{0}, N, A$}
    \STATE $H \gets \phi$
    \FOR {$n \gets 1$ to $N$}                   
        \STATE $x^* \gets max_{x}$ $A(x, M_{n-1})$,
        \STATE evaluate $f(x^*)$
        \STATE $H \gets H \cup (x^*, f(x^*))$,
        \STATE Fit a new model $M_{n}$ to $H$
    \ENDFOR
    \STATE \RETURN {$H$}
\eindent
\STATE \textbf{end function}
\end{algorithmic}
\end{algorithm}
\vspace*{-\baselineskip}
\noindent
Where H is the observation history of the pair (feature, value), N is the max number of iterations, f is the true objective function (black box), M is the surrogate function (Gaussian process), A is the Acquisition function, $x^*$ is the next feature chosen to evaluate~\cite{bo_pseudo_code}.

\subsubsection{Surrogate Model (Gaussian Process)} It uses the model's history to construct multivariate Gaussian distributions to predict the objective function. The \textit{mean} and \textit{covariance function} decide where to steer the acquisition function. %The \textbf{mean function} is most simply given by:

%\begin{equation} 
%\mu(x)
%= E[f(x)]
%\end{equation}
%\mbox{}\\
%Where $\mu(x)$ is the mean function at $x$ and $E[f(x)]$ represents the expected value of the objective function at $x$~\cite{gp_tutorial}. 
In our paper, we use the \textit{matern (covariance) kernel function} along with this. Matern kernels are a generalization of power exponential kernels where the additional parameter $\nu$ controls the smoothness of the function. The larger $\nu$ is the smoother the function until $\nu \rightarrow \infty$, where the matern kernel becomes a power exponential kernel~\cite{bayes_opt_info}. A 2.5 matern kernel has $\nu = 2.5$ for twice differentiable functions. This kernel is given by:

\begin{equation}
k(x_i, x_j) =  \frac{1}{\Gamma(\nu)2^{\nu-1}}\Bigg(
\frac{\sqrt{2\nu}}{l} d(x_i , x_j )
\Bigg)^\nu K_\nu\Bigg(
\frac{\sqrt{2\nu}}{l} d(x_i , x_j )\Bigg),
\end{equation}

\noindent
where $k$ is the kernel, $d(x_i , x_j)$ is the euclidean distance between $x_i$ and $x_j$, $\nu$ is the smoothness of the function, $K_\nu()$ is the modified Bessel function, and $\Gamma(\nu)$ is the gamma function~\cite{gp_tutorial}.

\subsubsection{Acquisition Function} This is optimized to balance the exploration and exploitation of the objective function. The \textbf{Upper Confidence Bound} is simple, and what we use given by:

\begin{equation}
a(x;\lambda) = \mu(x) + \lambda \sigma (x),
\end{equation}

\noindent
where $\mu(x)$ is the expected performance of our surrogate model, controlling the exploration, and $\sigma=\sqrt{k(x_i,x_j}$ adds uncertainty through the standard deviation (root of the kernel), controlling the exploitation~\cite{gp_tutorial}. The larger $\lambda$ is, the more this function is rewarded for searching unknown spaces, leading to more exploration~\cite{bo_tutorial}. 

\subsection{Relevance} 

The concept of relevance is defined by Ribeiro and Moniz~\cite{relevance_nuno} as a function given by:

\begin{equation}
\phi (Y):{\mathcal {Y}}\rightarrow [0,1],
\end{equation}

\noindent
where $\phi$ is the continuous relevance function, displaying the bias of domain $\mathcal {Y}$ by mapping it to the continuous range $[0, 1]$ with 0 being the minimum and 1 being the maximum relevance~\cite{relevance_def}. To calculate the relevance of all points, we require:

\begin{enumerate}
  \item[a)] \textbf{\textit{control points}} which are points with a known relevance
  \item[b)] \textbf{\textit{interpolation function}} that can determine the relevance of other points~\cite{relevance_nuno}.
\end{enumerate}

Control points are required for interpolation, with a minimum of three: the minimum, median, and maximum points for relevance. The control points can be defined by the equation given by:

\begin{equation}
S = \left\{ \langle y_k, \varphi (y_k), \varphi '(y_k) \rangle \right\} _{k=1}^s,
\end{equation}

\noindent
where $y_k$ is the control point, $\varphi(y_k)$ is the relevance at $y_k$, and $\varphi'(y_k)$ is the derivative of the relevance function at that point. By default, $\varphi'(y_k)$ will be 0 because the control points are assumed to be the minimums and maximums of the function~\cite{relevance_nuno}. Automatic generation of a relevance function makes common values less relevant and rarer points more relevant. We can also manually set the relevance of control points to reflect their real-world significance. Concerning interpolation, \cite{relevance_nuno} proposes the use of \textit{pchip} (Piecewise Cubic Hermite Interpolating Polynomials) to achieve the desirable conditions, such as continuity in generated points.

\section{Our Framework: Relevance-Aware Algorithmic Recourse}
In this section, we detail our framework, Relevance-Aware Algorithmic Recourse (RAAR). First, we introduce background information and details about our algorithmic recourse problem. Then, we introduce our objective function and how to optimize it.

\subsection{Preliminaries}
Through Bayesian optimization, we optimize for a particular outcome by applying changes to $x$, where $f(x+\epsilon)$ is the new outcome with a cost of $\epsilon$. We explore the solution space through the Gaussian process, generating a counterfactual distribution. Formally, we have a regression model $f: \chi \rightarrow {\rm I\!R}$ with the Gaussian Process surrogate Model, $GP(\mu_n, K_n)$. This surrogate model accepts a mean function $\mu_n : \chi \rightarrow {\rm I\!R}$ and a covariance function $K_n : \chi * \chi \rightarrow {\rm I\!R}$, with an Upper Confidence Bound acquisition function~\cite{bayes_opt_info, bo_alg_rec}. 

\subsection{Formulating and Optimizing our Objective}
We have two types of objective functions, one for $y$, the target value, and one for relevance, $\phi()$. We also have two optimization goals, one for obtaining the maximum $y$ value possible and one for a specific target value $y$. Additionally, we introduce bounds to either focus or expand their exploration. For maximum value optimizations, we set bounds of $\pm$5\% to focus on recourses with small changes. For target value optimizations, we use $\pm100\%$ to focus on reaching the target value in both y and relevance optimizations. This process compares the impact of relevance on recourses.

\subsubsection{Y Optimization} For this, the maximum value optimization uses the objective function as follows:
\begin{equation}
x^*=
\max_{x \epsilon X} \frac{f(x^*) - M(x)}{|M(x)|}
\end{equation}
For the target optimization, we optimize for the point that achieves the closest value to our target value.
\begin{equation}
x^*=
\max_{x \epsilon X} \frac{-|f(x^*) - y_{target}|}{|y_{target} - y_{orig}|} 
\end{equation}

\subsubsection{Relevance Optimization} We optimize for relevance through using $\phi(f(x))$ instead of $f(x)$, where $\phi(y)$ is the relevance function. The relevance function further pseudo-bounds the optimization process to provide a realistic and relevance-bound recourse. We use the same equation for both maximum and target optimizations.
\begin{equation}
x^*=
\max_{x \epsilon X} \phi(f(x^*))
\end{equation}
The above $\phi$ function can be generated in two main ways: changing whether we optimize for a maximum relevance value or a target y value. We generate an automated (distribution-based) relevance function with only the right extreme for the maximum relevance value. We use the training set to optimize for its most extreme values, investigating how using the relevance function for optimization compares with target value optimizations. For target values, we generate a unique relevance function with both extremes for every target value we want to reach (local). To define a unique relevance function for a target value, we define its control points as follows:
\begin{equation}
S = \left\{ \langle y_{min}, 0, 0 \rangle, \langle y_{original}, 0.5, 0 \rangle, \langle y_{target}, 1, 0 \rangle, \langle y_{max}, 0, 0 \rangle,  \right\}
\end{equation}
Setting these control points as above ensures that the optimization direction is always at maximum relevance or $y_{target}$.

\section{Experiments}
This paper aims to analyze if Relevance-Aware Algorithmic Recourse (RAAR) generates better counterfactuals. Here, we discuss the experimental process of evaluating our novel framework. We investigate through two methods of generating a recourse, with two different optimization objectives and two regression models each. The following describes how we leverage the well-known learning algorithms Random Forest and LGBM in our optimization~\cite{rf, lgbm}.

\begin{figure}[!ht]
    \centering
    \includegraphics[width=0.7\linewidth]{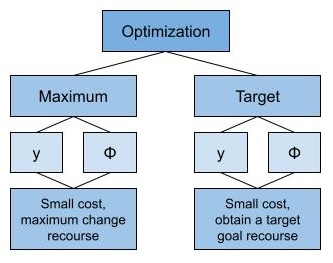}
    \caption{Experimental Process}
    \label{fig:image1}
\end{figure}

\subsection{Experimental Setup}
We compiled 15 datasets to run the process above, ranging from insurance expenses and mortgages to wine ratings. During every optimization process, we use Bayesian optimization to reach our target. For (1), the target is the maximum positive difference we can reach while minimizing the cost. For (2), the target is any value we provide, and we used four different percentage increases to analyze how close we can get to that target value. The results of these experiments answer how relevance affects recourses and how the two methods compare.

\subsection{Maximum Optimization Results} 
For the maximum optimization, we can break this method down further into two more: 1) context-free, which finds the maximum y, and 2) contextual, which finds the maximum relevance. By optimizing for a maximum y, we try to get as large (within bounds of $\pm$5\%) as possible without looking at the practical and real-world settings. However, by incorporating relevance, we introduce a further pseudo-bound by trying to get the largest y within the person's situation, providing a more realistic recourse.
\begin{table}[!ht]
\renewcommand{\arraystretch}{1.15}
    \scriptsize
    \centering
    \caption{Maximum Optimization Data Results—}\label{tab1}
    \begin{tabular}{|c|c|c|c|c|c|c|}
    \hline
        Model & Optimizes & $\Delta y\%$ Mean & $\Delta y\%$ SD & $\phi$ Mean & $\phi$ SD & $\Delta d$ \\
        \hline
        RF & Y & 30.2 & 61.74 & 0.18 & 0.23 & -0.07\% \\ 
        \cline{2-6}
         & Relevance & 25.36 & 60.98 & 0.18 & 0.23 & \\ \hline
        LGBM & Y & 41.23 & 65.37 & 0.24 & 0.26 & -0.12\% \\ \cline{2-6}
         & Relevance & 36.62 & 63.8 & 0.24 & 0.26 & \\ \hline

        % Model & Optimizes & $\Delta y\%$ Mean & $\Delta y\%$ SD & $\phi$ Mean & $\phi$ SD & $\Delta d$ \\
        % \hline
        % RF & Y & 26.98 & 57.75 & 0.19 & 0.23 & -0.1\% \\ 
        % \cline{2-6}
        %  & Relevance & 23.14 & 57.8 & 0.19 & 0.23 & \\ \hline
        % LGBM & Y & 35.71 & 57.22 & 0.25 & 0.26 & -0.15\% \\ \cline{2-6}
        %  & Relevance & 29.5 & 56.08 & 0.25 & 0.26 & \\ \hline

        % RF & Relevance & 23.14 & 57.8 & 0.19 & 0.23 & -0.1 \\ \hline
        % LGBM & Y & 35.71 & 57.22 & 0.25 & 0.26 & -0.15 \\ \hline
        % LGBM & Relevance & 29.5 & 56.08 & 0.25 & 0.26 & -0.15 \\ \hline
    \end{tabular}
\end{table}
\mbox{}\\
In Table 1, $\Delta y\%$ is the average percent change in y, with its standard deviation, $\phi$ mean and SD are the mean and standard deviation relevance values, $\Delta d$ is the distance difference between the y and relevance optimization models w.r.t. relevance. Recourses that leverage relevance have smaller cost (distance) and smaller y percent change. Here, our goal is to evaluate the recourses generated by the y-optimized and relevance-optimized models. Overall, the y optimization model recourses obtain a larger change in the outcome than relevance optimization models. However, they both reach the same relevance, outlining that the relevance functions produced limit the change in y. This is because we automatically generate it to favor larger, rarer extreme cases without guidance.

\subsection{Target Optimization Results}
In many scenarios, people want to know what they can change to achieve a specific outcome instead of a maximum. To replicate this, we optimized for a target value instead of a maximum value. We compared the y difference model against the relevance model when predicting values 10\%, 20\%, 50\%, and 100\% greater than its original. For the relevance model, we manually created relevance functions.

\begin{table}[!ht]
    \scriptsize
    \centering
    % explain what the labels are in caption
    \caption{Target Optimization Data Results}\label{tab1}
    \begin{tabular}{|c|c|c|c|c|c|c|c|c|}
    \hline
        Model & Target $\Delta y\%$ & Optimizes & $\Delta y\% $  & Y SD  & $\phi$ Mean & $\phi$ SD & $\Delta It$  & $\Delta d$  \\ \hline
        ~ & 10\% & Y & 9.57\% & 1.25 & 0.94 & 0.13 & -6.24\% & -2.15\% \\ \cline{3-7} 
        ~ & ~ & Relevance & 13.14\% & 1.52 & 0.96 & 0.10 & &  \\ \cline{2-9} 
        ~ & 20\% & Y & 18.02\% & 0.74 & 0.90 & 0.17 & -3.48\% & -1.13\% \\ \cline{3-7}
        RF & ~ & Relevance & 20.78\% & 0.89 & 0.93 & 0.13 &  & \\ \cline{2-9}
        ~ & 50\% & Y & 41.39\% & 0.41 & 0.87 & 0.18 & -1.19\% & -0.33\% \\ \cline{3-7}
        ~ & ~ & Relevance & 42.63\% & 0.48 & 0.90 & 0.13 &  &  \\ \cline{2-9}
        ~ & 100\% & Y & 55.87\% & 0.27 & 0.65 & 0.19 & -5.71\% & -0.82\% \\ \cline{3-7}
        ~ & ~ & Relevance & 48.83\% & 0.31 & 0.74 & 0.12 &  &  \\ \hline
         ~ & 10\% & Y & 9.61\% & 0.44 & 0.96 & 0.12 & -4.98\% & -1.91\% \\ \cline{3-7}
        ~ & ~ & Relevance & 13.09\% & 0.82 & 0.98 & 0.08 &  &  \\ \cline{2-9}
        ~ & 20\% & Y & 18.71\% & 0.27 & 0.94 & 0.13 & -5.89\% & -2.17\% \\ \cline{3-7}
        LGBM & ~ & Relevance & 21.08\% & 0.50 & 0.96 & 0.10 & &  \\ \cline{2-9}
        ~ & 50\% & Y & 44.03\% & 0.22 & 0.90 & 0.16 & -6.15\% & -1.76\% \\ \cline{3-7}
        ~ & ~ & Relevance & 43.40\% & 0.32 & 0.91 & 0.12 & & \\ \cline{2-9}
        ~ & 100\% & Y & 64.12\% & 0.2 & 0.66 & 0.18 & -8.38\% & -2.04\% \\ \cline{3-7}
        ~ & ~ & Relevance & 53.32\% & 0.24 & 0.76 & 0.12 &  &  \\ \hline
    \end{tabular}

% \begin{tabular}{|c|c|c|c|c|c|c|c|c|}
%     \hline
%         Model & Target $\Delta y\%$ & Optimizes & $\Delta y\% $  & Y SD  & $\phi$ Mean & $\phi$ SD & $\Delta It$  & $\Delta d$  \\ \hline
%         ~ & 10\% & Y & 9.56\% & 1.24 & 0.93 & 0.14 & -6.17\% & -1.94\% \\ \cline{3-7} 
%         ~ & ~ & Relevance & 13.83\% & 1.56 & 0.96 & 0.10 & &  \\ \cline{2-9} 
%         ~ & 20\% & Y & 18.12\% & 0.73 & 0.90 & 0.16 & -3.48\% & -0.93\% \\ \cline{3-7}
%         RF & ~ & Relevance & 21.52\% & 0.90 & 0.93 & 0.12 &  & \\ \cline{2-9}
%         ~ & 50\% & Y & 42.00\% & 0.39 & 0.88 & 0.17 & -1.15\% & -0.31\% \\ \cline{3-7}
%         ~ & ~ & Relevance & 43.87\% & 0.48 & 0.90 & 0.13 &  &  \\ \cline{2-9}
%         ~ & 100\% & Y & 59.19\% & 0.26 & 0.68 & 0.18 & -5.14\% & -0.88\% \\ \cline{3-7}
%         ~ & ~ & Relevance & 53.35\% & 0.30 & 0.76 & 0.12 &  &  \\ \hline
%          ~ & 10\% & Y & 9.63\% & 0.44 & 0.96 & 0.12 & -4.97\% & -1.67\% \\ \cline{3-7}
%         ~ & ~ & Relevance & 13.54\% & 0.86 & 0.98 & 0.07 &  &  \\ \cline{2-9}
%         ~ & 20\% & Y & 18.78\% & 0.27 & 0.94 & 0.12 & -5.74\% & -1.90\% \\ \cline{3-7}
%         LGBM & ~ & Relevance & 21.61\% & 0.51 & 0.96 & 0.10 & &  \\ \cline{2-9}
%         ~ & 50\% & Y & 44.44\% & 0.21 & 0.91 & 0.15 & -5.78\% & -1.47\% \\ \cline{3-7}
%         ~ & ~ & Relevance & 44.45\% & 0.32 & 0.92 & 0.11 & & \\ \cline{2-9}
%         ~ & 100\% & Y & 66.81\% & 0.19 & 0.68 & 0.17 & -7.61\% & -1.84\% \\ \cline{3-7}
%         ~ & ~ & Relevance & 57.47\% & 0.23 & 0.78 & 0.11 &  &  \\ \hline
%     \end{tabular}

\end{table}

Table 2: Target $\Delta y\%$ is the target percent change, $\Delta y\%$ is the average percent change in y, with its standard deviation, $\phi$ mean and SD are the mean and standard deviation relevance values, $\Delta It$ is the difference in number iterations to reach their recourse w.r.t. relevance, $\Delta d$ is the distance difference between the y and relevance optimization models w.r.t. relevance. Recourses that leverage relevance are quicker and cheaper, with outcomes similar to target optimization recourses.
\begin{figure}[!ht]
    \centering
    \includegraphics[width=\linewidth]{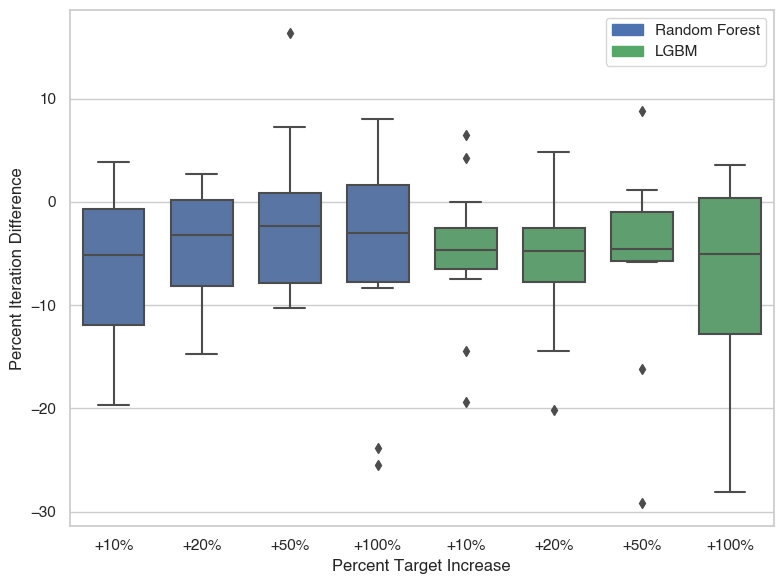}
    \caption{Percentage Optimization Iteration Difference}
    \label{fig:image1}
    \Description{Box plot showcasing the optimization iteration difference.}
    
\end{figure}
\begin{figure}[!ht]
    \centering
    \includegraphics[width=\linewidth]{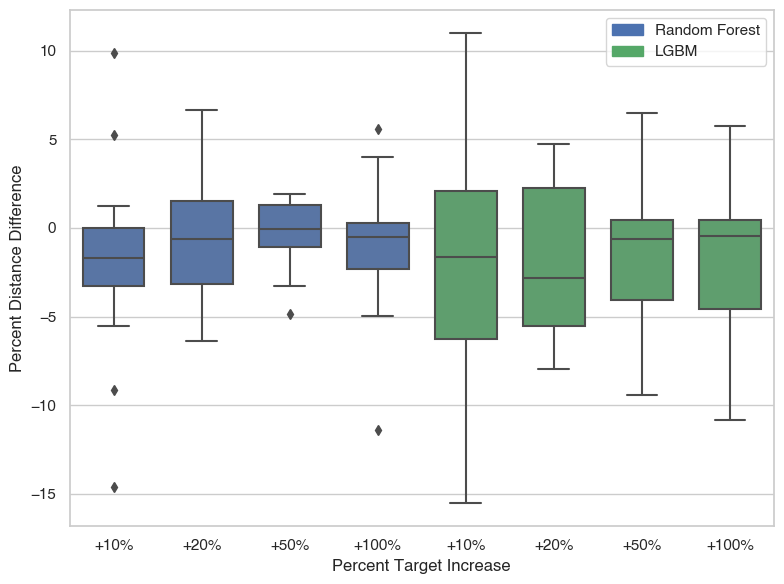}
        \caption{Percentage Optimization Distance Difference}
        \label{fig:image2}
    \Description{Box plot showcasing the optimization distance difference.}
\end{figure}

% \begin{figure}[!ht]
%     \centering
%     \begin{subfigure}{.245\textwidth}
%         \centering
%         \includegraphics[width=0.98\linewidth]{iter_diffs.png}
%         \caption{\% Iteration Difference}
%         \label{fig:image1}
%     \end{subfigure}%
%     \begin{subfigure}{.245\textwidth}
%         \centering
%         \includegraphics[width=0.98\linewidth]{dist_diffs.png}
%         \caption{\% Distance Difference}
%         \label{fig:image2}
%     \end{subfigure}
%     \caption{Optimization Iteration and Distance Comparisons}\label{fig:images}
% \end{figure}

% \begin{figure}
%     \caption{Optimization Iteration and Distance Comparisons}
%     \label{fig:images}
%     \centering
%     \includegraphics[width=.9\linewidth]{iter_diffs.png}
%     \caption{\% Iteration Difference}
%     \label{fig:image1}
% \end{figure}

% \begin{figure}
%     \centering
%     \includegraphics[width=.9\linewidth]{dist_diffs.png}
%     \caption{\% Distance Difference}
%     \label{fig:image2}
% \end{figure}
Here, we evaluate how the two optimization processes compare for a target value. In the table, the y optimization model finds recourses similar to the target on average only 60\% of the time. However, Table~\ref{tab1}, Figure~\ref{fig:image1}, and Figure~\ref{fig:image2} show that the iterations and distances for relevance optimization models find it quicker and cheaper, as demonstrated by the negative percentage. Moreover, LGBM provides quicker and closer recourses to the target value than random forest regression. Overall, relevance optimizations achieve results similar to y optimizations but with greater efficiency because of the reduced optimization iterations required and lower costs since the recourses generated are closer to the original points. This introduces a trade-off between greater efficiency versus a small reduction in accuracy and precision. 

\section{Conclusion \& Future Work}
In this paper, we proposed the novel framework for algorithmic recourse (RAAR), to address the issue of treating all values equally which is unrealistic in real-world settings. We introduced the notion of relevance that assigns values with varying levels of importance, which the model optimizes for. Through Bayesian optimization, we used objective functions, a Gaussian process, and an acquisition function to generate recourses that optimize for the maximum change with the smallest cost or target change. The 15 datasets and results show that relevance recourses are quicker and more cost-efficient than its y recourse methods. Furthermore, they can be similar sometimes, making relevance-focused recourses more desirable in certain situations. Our work also highlights future development on relevance-focused recourses. For instance, how different shapes of a relevance function can improve the accuracy and precision of such recourses or how different mean covariance and acquisition functions influence recourse. Additionally, using relevance-focused recourses with other approaches to recourse mentioned earlier, such as integer programming, would be useful to understand how relevance can aid in providing effective recourses.

\bibliographystyle{ACM-Reference-Format}
\bibliography{bibliography}

%\appendix

%\section{Appendix}


%Lorem ipsum dolor sit amet, consectetur adipiscing elit. Morbimalesuada, quam in pulvinar varius, metus nunc fermentum urna, id sollicitudin purus odio sit amet enim. Aliquam ullamcorper eu ipsum vel mollis. Curabitur quis dictum nisl. Phasellus vel semper risus, et lacinia dolor. Integer ultricies commodo sem nec semper.


%Etiam commodo feugiat nisl pulvinar pellentesque. Etiam auctor sodales ligula, non varius nibh pulvinar semper. Suspendisse nec lectus non ipsum convallis congue hendrerit vitae sapien. Donec at laoreet eros. Vivamus non purus placerat, scelerisque diam eu, cursus ante. Etiam aliquam tortor auctor efficitur mattis.


%Nam id fermentum dui. Suspendisse sagittis tortor a nulla mollis, in pulvinar ex pretium. Sed interdum orci quis metus euismod, et sagittis enim maximus. Vestibulum gravida massa ut felis suscipit congue. Quisque mattis elit a risus ultrices commodo venenatis eget dui. Etiam sagittis eleifend elementum.

%Nam interdum magna at lectus dignissim, ac dignissim lorem rhoncus. Maecenas eu arcu ac neque placerat aliquam. Nunc pulvinar massa et mattis lacinia.

\end{document}